\documentclass[runningheads]{llncs}
\usepackage{graphicx}
\usepackage{comment}
\usepackage{amsmath,amssymb} % define this before the line numbering.
\usepackage{color}
\usepackage{booktabs}
\usepackage{makecell}
\usepackage{multirow}
\newcommand{\tabincell}[2]{\begin{tabular}{@{}#1@{}}#2\end{tabular}}

\usepackage[width=122mm,left=12mm,paperwidth=146mm,height=193mm,top=12mm,paperheight=217mm]{geometry}

\begin{document}
% \linenumbers
\pagestyle{headings}
\mainmatter

\title{Learning Generalized Spoof Cues for Face Anti-spoofing} % Replace with your title

%\begin{}
%\titlerunning{ECCV-20 submission ID \ECCVSubNumber} 
\titlerunning{ }
%\authorrunning{ECCV-20 submission ID \ECCVSubNumber} 
\authorrunning{ }
%\author{Anonymous ECCV submission}
\author{Haocheng Feng\inst{1} \and Zhibin Hong\inst{1} \and Haixiao Yue\inst{1} \and Yang Chen\inst{2 \footnotemark[1]} \and Keyao Wang\inst{1} \and Junyu Han\inst{1} \and Jingtuo Liu\inst{1} \and Errui Ding\inst{1}}
\institute{Department of Computer Vision Technology(VIS), Baidu Inc.\inst{1}
\\Beihang University\inst{2}\\
\email{\{fenghaocheng, hongzhibin, yuehaixiao, wangkeyao,\\hanjunyu, liujingtuo, dingerrui\}@baidu.com \\ chenyangwiz@buaa.edu.cn}}

%\end{comment}

%******************

\begin{comment}
\titlerunning{Abbreviated paper title}
\author{Feng Haocheng\inst{1}\orcidID{0000-1111-2222-3333} \and
Second Author\inst{2,3}\orcidID{1111-2222-3333-4444} \and
Third Author\inst{3}\orcidID{2222--3333-4444-5555}}
%
\authorrunning{F. Author et al.}
\institute{Department of Computer Vision Technology(VIS), Baidu Inc.
\email{lncs@springer.com}\\
\url{http://www.springer.com/gp/computer-science/lncs} \and
ABC Institute, Rupert-Karls-University Heidelberg, Heidelberg, Germany\\
\email{\{abc,lncs\}@uni-heidelberg.de}}
\end{comment}
%******************
\maketitle

\renewcommand{\thefootnote}{\fnsymbol{footnote}}
\footnotetext[1] {This work was done when Yang Chen was an intern at Baidu Inc. }

\begin{abstract}
Many existing face anti-spoofing (FAS) methods focus on modeling the decision boundaries for some predefined spoof types. However, the diversity of the spoof samples including the unknown ones hinders the effective decision boundary modeling and leads to weak generalization capability. In this paper, we reformulate FAS in an anomaly detection perspective and propose a residual-learning framework to learn the discriminative live-spoof differences which are defined as the spoof cues. The proposed framework consists of a spoof cue generator and an auxiliary classifier. The generator minimizes the spoof cues of live samples while imposes no explicit constraint on those of spoof samples to generalize well to unseen attacks. In this way, anomaly detection is implicitly used to guide spoof cue generation, leading to discriminative feature learning. The auxiliary classifier serves as a spoof cue amplifier and makes the spoof cues more discriminative. We conduct extensive experiments and the experimental results show the proposed method consistently outperforms the state-of-the-art methods. The code will be publicly available at \url{https://github.com/vis-var/lgsc-for-fas}.

\end{abstract}

\section{Introduction}
Face authentication applications have been widely used in our daily lives. However, hackers may fraud the face recognition systems by the means of presentation attack (PA) such as printed photos (\textit{i.e.} print attack), digital images / videos (\textit{i.e.} replay attack) and 3D facial masks (\textit{i.e.} 3D mask attack). To this end, face anti-spoofing (FAS) is developed to guarantee the security of the face recognition system. 

Traditional methods employ hand-crafted features such as LBP~\cite{maatta2011face}, HOG \cite{komulainen2013context} and SIFT \cite{patel2016secure} to extract the texture information and use shallow classifiers to model the decision boundary. The hand-crafted features used by such methods are not discriminative enough and the classifier’s performance is limited. To take advantage of better feature representations, the early deep learning based methods \cite{xu2015learning,yang2014learn} utilize CNN to learn more discriminative features and formulate the spoofing detection as a binary classification problem. However, these methods tend to overfit on the predefined datasets and cannot generalize well. Most recent deep learning based methods \cite{liu2018learning,qin2019learning,shao2019regularized} focus on improving the model's generalization capacity and achieve promising results.

The limited generalization capacity of FAS lies in the diversity of the spoof samples including the unknown ones. Though one can assume that the live samples share the same nature and can be categorized into one class, the spoof samples are generally diverse due to the wide variety of attack mediums. To relieve the spoof diversity's influence on the decision boundary modeling, we reformulate FAS from an anomaly detection perspective and assume the live samples belong to a closed-set while the spoof ones are outliers of the closed-set and belong to an open-set. Based on the hypothesis, we define the spoof cues as the discriminative features that can be used to differentiate the closed-set and the open-set.

\begin{figure}[t]
\centering
\includegraphics[width=29pc, height=9pc]{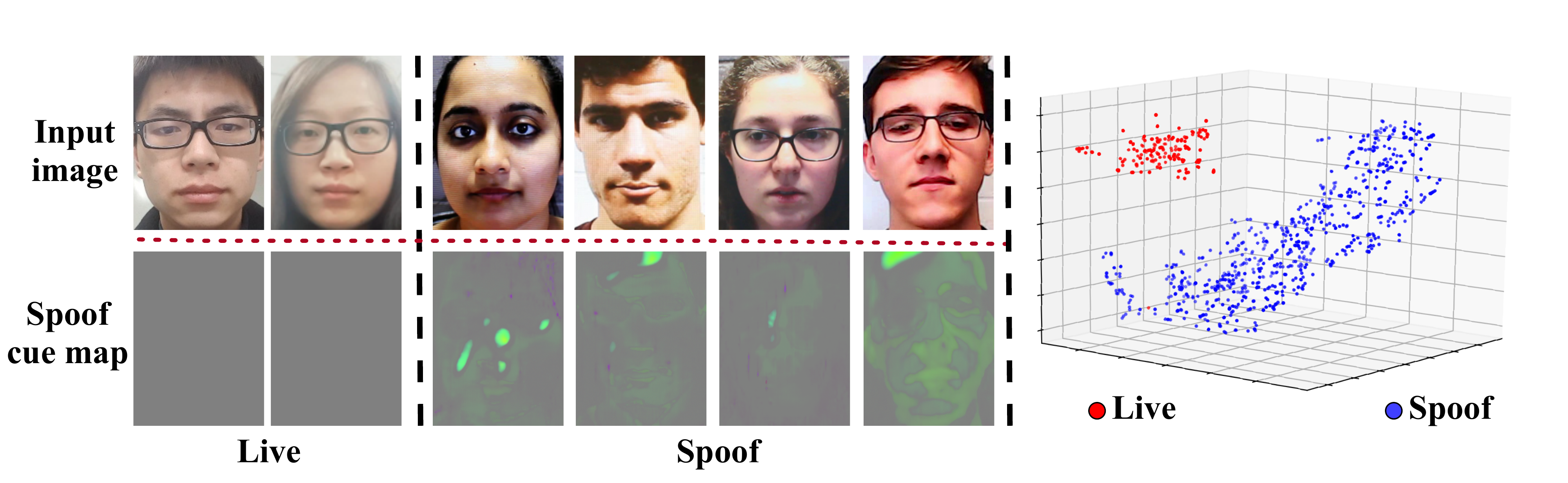}
\caption{The illustration of the spoof cue maps and the distribution of the learned feature embeddings. \textbf{Left}: the input images and the generated spoof cue maps. \textbf{Right}: the distribution of the learned feature embeddings visualized using the t-SNE. More details will be given in Section \ref{ablations}.}
\label{scan}
\end{figure}

In this paper, the spoof cues are formulated as a feature map that has the same size as the input image. The spoof cue map is prone to be a non-zero map for a spoof image while an all-zero map for a live image. Unlike the compact embeddings which are traditionally used for classification (live vs. spoof), spoof cues can effectively encode the spatial information. To learn the spoof cue map, we propose the residual-learning framework that consists of a spoof cue generator and an auxiliary classifier. In the spoof cue generator, we set explicit regression loss %on the generator's output 
for live samples to minimize the magnitudes of their spoof cues while setting no explicit constraint on the spoof samples, leading their corresponding entries to any real numbers. In this way, the spoof cues of live and spoof samples are naturally separable. In addition,
%setting no explicit constraint for the outputs of the spoof ones. Additionally, 
we adopt the multi-scale feature-level metric learning on the generator to promote live-live intra-class compactness and live-spoof inter-class separability. % as the implicit supervision.
Then, the spoof cue map and the input image, which are skip-connected in a residual learning manner, are fed to an auxiliary classifier, further improving the discriminative feature learning.
%Overall, We feed the spoof cue map and the input image to an auxiliary classifier in a residual learning manner which further amplifies the spoof cues and makes the spoof cues more discriminative.
The spoof cue maps and their distributions are shown in Fig.~\ref{scan}. The spoof cue maps are directly utilized to calculate the score of an input sample being spoof in the test phase. The experimental results show that the spoof cue maps can be utilized to separate the live closed-set and spoof open-set effectively.

%To sum up, our contributions can be concluded as follows:
Our contributions can be summarized as:
\begin{enumerate}
	\item We reformulate FAS from an anomaly detection perspective and propose a spoof cue generator, which models the spoof cues of live samples on closed-set and spoof samples on open-set by minimizing the magnitudes of spoof cues for live samples while imposing no explicit constraint on the spoof samples.
	%with weak constraints on the spoof samples to learn the generalized spoof cues separating the live closed-set and the spoof open-set.
	\item We deploy a residual learning framework to make the spoof cues more discriminative and further boost the spoof cues' generalization capability.
	\item Without additional depth or temporal information, the proposed framework outperforms the state-of-the-art on the popular RGB anti-spoofing datasets.
\end{enumerate}

\section{Related Work}

\subsection{Face anti-spoofing methods}
The traditional face anti-spoofing methods use the hand-crafted features, such as LBP \cite{maatta2011face}, HOG \cite{komulainen2013context} and SIFT \cite{patel2016secure} to represent the image texture and use the traditional classifiers like SVM for binary classification. Besides, some works also adopt the feature extractors in the different color spaces like HSV and YCbCr \cite{boulkenafet2015face} to tackle face anti-spoofing. However, the hand-crafted features are specially designed for several defined types of spoof and are not robust enough to variable conditions such as the illumination and the presentation attack types. Motion cues like eye-blinking \cite{pan2007eyeblink} and head movement \cite{kollreider2007real} are also used when the temporal information is given. However, the performance of these methods drop drastically on the video replay attack.

With the development of CNN, many researchers start to use CNN as a feature extractor to exploit the discriminative representations in RGB images and formulate face anti-spoofing as a binary classification problem \cite{xu2015learning}.
Moreover, Feng \emph{et~al.} \cite{feng2016integration} utilize LSTM in temporal information modeling for better classification performance. Despite their good performances in the intra-dataset test, these methods are easy to overfit and have poor generalizability.

Most recent methods can be categorized into three classes: the additional information-based methods, the domain shift methods, and the few-shot methods. These methods are designed to further boost the models' generalizability. Liu \emph{et~al.} \cite{liu2018learning} use pseudo-depth to supervise the model training. Shao \emph{et~al.} \cite{shao2019regularized} cast FAS as a domain generalization problem and develop a Regularized Fine-grained Meta-learning framework. Qin \emph{et~al.} \cite{qin2019learning} propose a novel Adaptive Inner-update Meta Face Anti-Spoofing (AIM-FAS) method.
\subsection{Anomaly detection}
Anomaly detection is the task to identify the unusual samples from a set of normal data. It has been widely used in a variety of applications such as fraud detection \cite{adewumi2017survey}, cyber-intrusion detection \cite{kwon2018empirical}, health care \cite{wang2016research} and video surveillance \cite{au2006anomaly}.
In the face anti-spoofing area, Liu \emph{et~al.} \cite{arashloo2017anomaly} formulate the face PAD problem as anomaly detection and put forward a one-class classifier. Nikisins \emph{et~al.} \cite{nikisins2018effectiveness} use a Gaussian Mixture model-based anomaly detector. Arashloo \emph{et~al.} \cite{arashloo2018client} use client-specific information in a one-class anomaly detection formulation to improve the model's performance significantly. Perez \emph{et~al.} \cite{perez2019deep} reformate the FAS problem from an anomaly detection perspective and put forward a deep metric learning model with triplet focal loss. 
\section{Our Approach}
\subsection{Preliminaries}
\noindent \textbf{Anomaly detection.}
\label{Anomaly detection}
Anomaly detection (AD) is the task of identifying the unusual samples in a set of normal data. The typical AD methods attempt to learn a compact description of the normal data in an unsupervised manner. Take the deep SVVD \cite{ruff2018deep} as an example, for input space $(\mathcal{X} \subseteq \mathbb{R}^{d})$ and output space $(\mathcal{Z} \subseteq \mathbb{R}^{p})$, let $\phi (\cdot ;\mathcal{W}) : \mathcal{X} \rightarrow \mathcal{Z}$ be a neural network with ${L}$ hidden layers and corresponding set of weights $\mathcal{W} = {\left \{ \mathcal{W}^{1},...,\mathcal{W}^{L} \right \}}$. The objective is to train the neural network $\phi$ to learn a transformation that minimizes the volume of a data-enclosing hypersphere in output space $\mathcal{Z}$ centered on a predetermined point $c$. Given $N$ (unlabeled) training samples $(x_1,...,x_N \subseteq \mathcal{X})$, the objective is:
\begin{equation}
\setlength{\abovedisplayskip}{3pt}
\setlength{\belowdisplayskip}{3pt}
\begin{aligned}
\underset{\mathcal{W}}{min } \frac{1}{N}\sum_{i=1}^{N}\left \| \phi (x_i;\mathcal{W} )- c \right \|^{2} + \frac{\lambda }{2} \sum_{{\ell}=1}^{L}\left \| \mathcal{W}^{\ell} \right \|_{F}^{2}, \lambda  > 0,
\end{aligned}
\end{equation}
Once the network is trained, the anomaly score for a test point $x$ is given by the distance from $\phi(x;\mathcal{W})$ to the center of the hypersphere: 
\begin{equation}
\setlength{\abovedisplayskip}{3pt}
\setlength{\belowdisplayskip}{3pt}
\begin{aligned}
s(x) = \left \| \phi (x;\mathcal{W})- c \right \|.
\end{aligned}
\end{equation}

In FAS, though the live samples are assumed to share the same nature, the spoof samples can be very diverse due to the wide variety of attack mediums. Such diversity makes it hard for the spoof samples to form a compact region in the feature representation space and further hinders the effective decision boundary modeling between live and spoof samples. Besides, the decision boundary learned by the known spoof samples may have poor performance on the unseen spoof samples.
Motivated by the AD method, we assume the live samples belong to a closed-set while the spoof samples are outliers from this closed-set and belong to an open-set. Therefore, the distribution of the live samples is assumed to lie in a compact sphere in the learned feature representation space while the spoof samples far away from the center of the live sphere. Specifically, for the face anti-spoofing model's input space $(\mathcal{X}\subseteq \mathbb{R}^{d})$ and output space $(\mathcal{Z} \subseteq \mathbb{R}^{p})$. Let $\phi (\cdot ;\mathcal{W}) : \mathcal{X} \rightarrow \mathcal{Z}$ be a neural network with ${L}$ hidden layers and corresponding set of weights $\mathcal{W} = {\left \{ \mathcal{W}^{1},...,\mathcal{W}^{L} \right \}}$. Given $N_{l}$ live samples $(x_1,...,x_{N_{l}} \subseteq \mathcal{X})$, $N_{s}$ spoof samples $(y_1,...,y_{N_{s}} \subseteq \mathcal{X})$, let $c$ be the center of the live samples in the output space $\mathcal{Z}$, the objective is:

\begin{equation}
\setlength{\abovedisplayskip}{3pt}
\setlength{\belowdisplayskip}{3pt}
\begin{aligned}
\underset{\mathcal{W}}{min } \frac{1}{N_{l}}\sum_{i=1}^{N_{l}}\left \| \phi (x_i;\mathcal{W} )- c \right \|^{2},
\label{obj1}
\end{aligned}
\end{equation}
\begin{equation}
\setlength{\abovedisplayskip}{3pt}
\setlength{\belowdisplayskip}{3pt}
\begin{aligned}
\underset{\mathcal{W}}{max } \frac{1}{N_{s}}\sum_{i=1}^{N_{s}}\left \| \phi (y_i;\mathcal{W} )- c \right \|^{2}.
\label{obj2}
\end{aligned}
\end{equation}

In the test phase, the score for a test sample $t$ being spoof can be given by the distance from $\phi (t;\mathcal{W})$ to the center of the live hypersphere:

\begin{equation}
\setlength{\abovedisplayskip}{3pt}
\setlength{\belowdisplayskip}{3pt}
\begin{aligned}
s(x) = \left \| \phi (t;\mathcal{W})- c \right \|.
\label{test_score}
\end{aligned}
\end{equation}

In our proposed method, we utilize the explict regression supervision on the live samples to achieve the optimization goal of Eq. \ref{obj1} and the implict metric learning supervision on the live and spoof samples to ahieve the optimazation goal of Eq. \ref{obj2}. In the test phase, we directly use the distance between the test samples and the predfined live closed set's center in the feature space to calculate the score of a sample being spoof.

\begin{figure}[t]
\centering
\includegraphics[width=28pc, height=9pc]{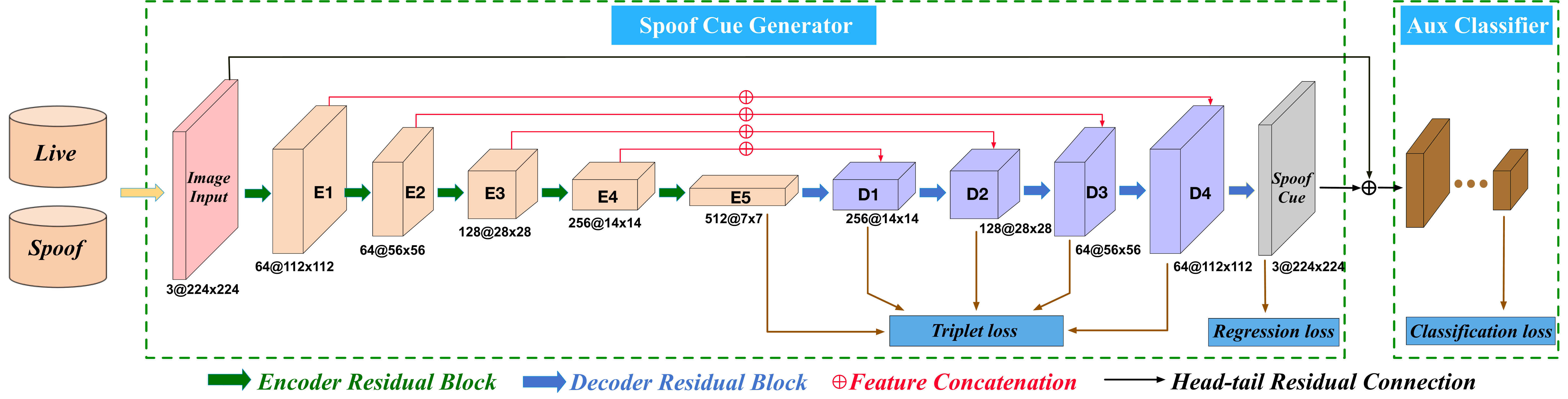}
\caption{The proposed network architecture.}
\label{fig_model}
\end{figure}

\noindent \textbf{Residual learning.}
\label{Residual learning}
Deep residual learning \cite{he2016deep} has gained great success in visual recognition, the residual learning method recasts the desired underlying mapping $H(x)$ into $H(x)=F(x)+x$ and learns the effective residual feature representation.

In FAS, the discriminative differences between the live and spoof samples may be caused by the intrinsic properties of the mediums such as printed papers or screens. Motivated by the residual learning method, the discriminative differences between live and spoof samples can be regarded as the residual of the spoof samples and the live samples. We define the discriminative differences as spoof cues and assume the spoof cues only exist in the spoof samples. In order to learn the spoof cues, we propose a residual learning framework that consists of a spoof cue generator and an auxiliary classifier. In the spoof cue generator, we impose the regression loss on the live samples while put no explicit supervision on the spoof samples, which guarantees the spoof cues' generalization capability. The auxiliary classifier serves as a spoof cue amplifier in the end-to-end training process and helps to learn more discriminative spoof cues. 
\subsection{Spoof cue learning framework}
The proposed residual learning framework is illustrated in Fig.~\ref{fig_model}. It consists of a spoof cue generator and an auxiliary classifier. In our assumption, the live samples share some common properties that spoof samples do not have. Therefore, all the live samples belong to a closed-set while the spoof samples are outliers from this closed-set but belong to an open-set. The discriminative differences between the live and spoof samples which are defined as spoof cues can be caused by the intrinsic properties of the medium carriers such as printed paper or screens. Our framework aims to learn the discriminative spoof cues and further boost the spoof cues' generalization capability.

In the spoof cue generator, we adopt a simple yet effective U-Net to generate spoof cue maps. In our assumption, the spoof cues only exist in spoof samples. Thus, for live samples, the spoof cue map should be an all-zero map. For spoof samples, the spoof cue map should be an unknown non-zero map. To relieve the spoof diversity’s influence on the decision boundary modeling and make the spoof cue generalized enough, we introduce the pixel-wise regression loss as explicit supervision only for the live samples. To promote the intra-class compactness in live samples and the inter-class separability between live and spoof samples, we introduce the multi-scale triplet loss on the feature embeddings. 

In the auxiliary classifier, we feed the spoof cue map and the input image into the auxiliary classifier in a residual learning manner. The auxiliary classifier serves as an amplifier in the end-to-end training process and makes the learned spoof cues more discriminative.

\subsection{Spoof Cue Generator}
\noindent \textbf{Architecture.}
\label{Architecture for Spoof Cue Mining}
In the spoof cue generator, as shown in Fig.~\ref{fig_model},  we adopt the U-Net architecture which builds skip connections from encoder to decoder in multiple scales to generate spoof cues. We choose the ResNet18 \cite{he2016deep} pre-trained on ImageNet \cite{deng2009imagenet} as encoder E which contains four Encoder Residual Blocks. Followed by E, the decoder D, composed of five Decoder Residual Blocks, decodes the information back to generate the spoof cues. In each Decoder Residual Block, the feature map from the previous layer is upsampled by a nearest-neighbor interpolation, following which we add a 2$\times$2 convolution. We aggregate the feature maps from the symmetric position in the encoder to the decoder by the concatenation operation and add an Encoder Residual Block afterward. At the end of the decoder, a Decoder Residual Block accompanied by a Tanh activation layer is applied to generate the outputs.

\noindent \textbf{Regression loss.}
\label{Regression loss for Spoof Cue Mining}
In our assumption, the live samples belong to a closed-set while the spoof samples are outliers from this closed-set and belong to an open-set. Therefore, we do not set any preset constraint on the spoof cues and only impose constraints on the live ones. Since we assume the spoof cues only exist in spoof samples, the spoof cue maps should be zero maps for live samples while remaining unknown for spoof samples. From our perspective, the zero map can be considered as the center of the live samples in the feature space and the live-only regression loss achieves the optimization goal of Eq. \ref{obj1}.

Given an RGB image $I$ 
%$I \in \mathbb{R}^{224\times224\times3}$
as input, the spoof cue generator outputs a spoof cue map of the same size, the spoof cue map $C$
%$C \in \mathbb{R}^{224\times224\times3}$
is a zero map for a live sample. While $C$ remains unknown for a spoof sample. The spoof cue regression loss for a live sample is the pixel-wise $L1$ loss, in the formulation of:
\begin{equation}
\setlength{\abovedisplayskip}{3pt}
\setlength{\belowdisplayskip}{3pt}
	L_r = \frac{1}{N_l} \sum_{I_i \in live} {||C_i||_1},
\end{equation}
\noindent where $N_l$ is the number of live samples in one batch.

\noindent \textbf{Metric Learning.}
\label{metric loss}
The metric learning-based loss is designed as an implicit supervision for spoof samples to promote live-live intra-class compactness and live-spoof inter-class separability on the feature level, which achieves the optimization goal of Eq. \ref{obj2}. Specifically, we obtain a set of feature vectors $\{V\}$ by employing the global average pooling (GAP) on feature maps from layer E5 to D4 and apply the metric supervision afterward. We use the triplet loss of which the anchor always belongs to the live class. The metric learning-based loss can be formulated as:
\begin{equation}
\setlength{\abovedisplayskip}{3pt}
\setlength{\belowdisplayskip}{3pt}
\begin{aligned}
	&L_t = \frac{1}{T} \sum_{i=1}^T max(d(a_i, p_i) - d(a_i, n_i) + m, 0), \\
	&d(i, j) =||\frac{v_i}{||v_i||_2} - \frac{v_j}{||v_j||_2}||_2,
\end{aligned}
\end{equation}

\noindent where $\{a_i, p_i, n_i\}$ denotes the anchor (live), positive (live), negative (spoof) samples within the $i$th triplet respectively, $T$ denotes the number of triplets, $d(i, j)$ represents the euclidean distance between two L2-normalized feature vectors output by the GAP layer, and $m$ is the pre-defined margin constant.

As for the triplet sampling strategy, we choose the online batch-all triplet mining proposed in~\cite{hermans2017defense}. At each training step, we collect all the valid triplets within the current batch of data for metric loss computation, the triplets satisfying $||d(a, n) - d(a, p)||_2 < m$. 
\subsection{Auxiliary Classifier}
\label{aux}
The success of prior work \cite{he2018triplet} proves the metric learning based loss in feature embeddings coupled with classification supervision helps to learn more discriminative features, which inspires us to design the auxiliary classifier. In our residual learning framework, the auxiliary classifier serves as a spoof cue amplifier and helps to learn more discriminative spoof cues. After the spoof cue generator, the generated spoof cue maps $C$ are added back to the input image $I$ to form the overlayed images $S$. With $S$ as input, the auxiliary classification loss can be formulated as:
\begin{equation}
\setlength{\abovedisplayskip}{3pt}
\setlength{\belowdisplayskip}{3pt}
	L_a = \frac{1}{N} \sum_{i=1}^{N} {z_i \log q_i + (1-z_i) \log (1-q_i)},
\end{equation}
\noindent where $N$ is the number of samples, $z_i$ is the binary label and $q_i$ is the network prediction. 

In our visualization experiments, we notice that choosing $S$ instead of $C$ as the input of the auxiliary classifier helps to learn more discriminative spoof cue maps, which proves the residual learning's superiority. We discuss the experimental details in Section \ref{ablations}.
\subsection{Training and Testing}
\noindent \textbf{Loss functions.} The loss functions of the proposed model are three-fold: the pixel-wise $L1$ loss $L_r$ for spoof cue regression on live samples, the triplet loss $L_t$ and the auxiliary binary classification loss $L_a$ on both live and spoof samples. We integrate all these losses and establish the total loss of $L$ during training:
\begin{equation}
\setlength{\abovedisplayskip}{3pt}
\setlength{\belowdisplayskip}{3pt}
	L = \alpha_1 L_r + \alpha_2\sum_{k\in{\{E5-D4\}}}{L_t^k} + \alpha_3 L_a,
\end{equation}
\noindent where $k$ indexes the layer where we apply the triplet loss, and $\alpha_1, \alpha_2, \alpha_3$ are the weights to balance the influence of the different loss components.

\noindent \textbf{Test Strategy.}
At the test stage, we use the generated spoof cue map instead of the classifier's output to evaluate. In Eq. \ref{test_score}, the distance from $\phi (t;\mathcal{W})$ to the center of the live hypersphere is utilized to calculate the spoof score. In our proposed method the all-zero map can be considered as the center of the live samples in the feature space. We directly use the spoof cue map to evaluate. We obtain the spoof cue map $\hat{C}$ and define the spoof score as the element-wise mean of $\hat{C}$ magnitude. The spoof score evaluates the probability of the test sample being spoof:
\begin{equation}
\label{score_equ}
\setlength{\abovedisplayskip}{3pt}
\setlength{\belowdisplayskip}{3pt}
	score = \overline{ ||\hat{C}||_1},
\end{equation}
\noindent The input image is likely to be a spoof one when the spoof score is high.

\begin{table}
\caption{The intra testing results and comparison with the state-of-the-art methods on 3 protocols of SiW dataset.}
\label{intra_siw}
\centering
\renewcommand\arraystretch{0.95}
 \setlength{\tabcolsep}{0.8mm}
\begin{tabular}{|c|c|c|c|c|}
	\hline
	  \small Protocol & \small Method & \small APCER (\%) & \small BPCER (\%) & \small ACER (\%) \\
	\hline
	\multirow{4}{*}{1} &
	  \small Auxiliary~\cite{liu2018learning} & \small 3.58 & \small 3.58 & \small 3.58 \\
	\cline{2-5}
	& \small STASN~\cite{yang2019face} & \small -- & \small -- & \small 1.00 \\
	\cline{2-5}
	& \small Meta-FAS-DR~\cite{zhao2019meta} & \small 0.52 & \small \textbf{0.50} & \small 0.51 \\
	\cline{2-5}
	& \small \textbf{Ours} & \small \textbf{0.00} & \small \textbf{0.50} & \small \textbf{0.25} \\
	\hline
	
	\multirow{4}{*}{2} &
	\small Auxiliary~\cite{liu2018learning} & \small 0.57$\pm$0.69 & \small 0.57$\pm$0.69 & \small 0.57$\pm$0.69 \\
	\cline{2-5}
	& \small Meta-FAS-DR~\cite{zhao2019meta} & \small 0.25$\pm$0.32 & \small 0.33$\pm$0.27 & \small 0.29$\pm$0.28 \\
	\cline{2-5}
	& \small STASN~\cite{yang2019face} & \small -- & \small -- & \small 0.28$\pm$0.05 \\
	\cline{2-5}
	& \small \textbf{Ours} & \small \textbf{0.00$\pm$0.00} & \small \textbf{0.00$\pm$0.00} & \small \textbf{0.00$\pm$0.00} \\
	\hline
	
	\multirow{4}{*}{3} &
	\small STASN~\cite{yang2019face} & \small -- & \small -- & \small 12.10$\pm$1.50 \\
	\cline{2-5}
	& \small Auxiliary~\cite{liu2018learning} & \small 8.31$\pm$3.81 & \small 8.31$\pm$3.80 & \small 8.31$\pm$3.81 \\
	\cline{2-5}
	& \small Meta-FAS-DR~\cite{zhao2019meta} & \small 7.98$\pm$4.98 & \small 7.35$\pm$5.67 & \small 7.66$\pm$5.32 \\
	\cline{2-5}
	& \small \textbf{Ours} & \small \textbf{1.61$\pm$1.69} & \small \textbf{0.77$\pm$1.09} & \small \textbf{1.19$\pm$1.39} \\
	\hline
\end{tabular}
\end{table}

\section{Experiments}
\subsection{Experiment Settings}
In our experiments, we crop the human faces out of the images and feed them to the proposed framework. For datasets that offer no face location ground truth, we use the Dlib \cite{king2009dlib} toolbox as the face detector. We resample the training examples to keep the live-spoof ratio to 1:1. The training batch size is 32 and the warm-up learning rate strategy is applied at the first training epoch. We choose Adam \cite{kingma2014adam} as the optimizer with initial learning rate 1e-3, which decays every 600 training steps by a factor of 0.95. In the spoof cue generator, we use Tanh as the activation function for the output layer of the U-Net, so the output distributes in the same range with the input image, which is linearly normalized to [-1, 1]. We set $m$ to 0.5 in the triplet loss and set $\alpha_1$ to $\alpha_3$ as 5, 1 and 5, respectively. The score threshold in Eq. \ref{score_equ} is set experimentally, and 0.01 is recommended. Our framework can be trained end-to-end and converges in 20 epochs on all the datasets used in the paper.

As for the metrics, we use Average Classification Error Rate (ACER) which is half of the summation of Attack Presentation Classification Error Rate (APCER) and Bona Fide Presentation Classification Error Rate (BPCER) for intra-dataset test on SiW \cite{liu2018learning} and OULU-NPU \cite{boulkenafet2017oulu} datasets:
\begin{equation}
\setlength{\abovedisplayskip}{3pt}
\setlength{\belowdisplayskip}{3pt}
	ACER = \frac{APCER + BPCER}{2}.
\end{equation}
Note that APCER corresponds to the highest false positive rate in all presentation attack instruments.

For inter-dataset test between CASIA-MFSD \cite{zhang2012face} and Replay-Attack \cite{chingovska2012effectiveness} datasets, we use the Half Total Error Rate (HTER) which is the mean of False Rejection Rate (FRR) and False Acceptance Rate (FAR):
 \begin{equation}
 \setlength{\abovedisplayskip}{3pt}
\setlength{\belowdisplayskip}{3pt}
 	HTER = \frac{FRR + FAR}{2}.
 \end{equation}

\begin{table}
\caption{The intra testing results and comparison with the state-of-the-art methods on 4 protocols of OULU-NPU dataset.}
\centering	
\renewcommand\arraystretch{0.9}
\setlength{\tabcolsep}{0.85mm}
\begin{tabular}{|c|c|c|c|c|}
	\hline
	\small Protocol & \small Method & \small APCER (\%) & \small BPCER (\%) & \small ACER (\%) \\
	\hline
	\multirow{5}{*}{1}
%	& \small GRADIANT~\cite{boulkenafet2017competition} & \small 1.3 & \small 12.5 & \small 6.9 \\
%	\cline{2-5}
	& \small MILHP~\cite{lin2018live} & \small 8.3 & \small 0.8 & \small 4.6 \\
	\cline{2-5}
	& \small STASN~\cite{yang2019face} & \small 1.2 & \small 2.5 & \small 1.9 \\
	\cline{2-5}
	& \small Auxiliary~\cite{liu2018learning} & \small 1.6 & \small 1.6 & \small 1.6 \\
	\cline{2-5}
	& \small FaceDs~\cite{jourabloo2018face} & \small 1.2 & \small 1.7 & \small 1.5 \\
	\cline{2-5}
	& \small \textbf{Ours} & \small \textbf{0.8} & \small \textbf{0.0} & \small \textbf{0.4} \\
	\cline{2-5}
	
	\hline
	
	\multirow{5}{*}{2}
%	& \small MILHP~\cite{lin2018live} & \small 5.6 & \small 5.3 & \small 5.4 \\
%	\cline{2-5}
	& \small FaceDs~\cite{jourabloo2018face} & \small 4.2 & \small 4.4 & \small 4.3 \\
	\cline{2-5}
	& \small Auxiliary~\cite{liu2018learning} & \small 2.7 & \small 2.7 & \small 2.7 \\
	\cline{2-5}
	& \small GRANDINT~\cite{boulkenafet2017competition} & \small 3.1 & \small 1.9 & \small 2.5 \\
	\cline{2-5}
	& \small STASN~\cite{yang2019face} & \small 4.2 & \small \textbf{0.3} & \small 2.2 \\
	\cline{2-5}
	& \small \textbf{Ours} & \small \textbf{0.8} & \small 0.6 & \small \textbf{0.7} \\
	
	\hline
	
	\multirow{5}{*}{3}
%	& \small MILHP~\cite{lin2018live} & \small \textbf{1.5$\pm$1.2} & \small 6.4$\pm$6.6 & \small 4.0$\pm$2.9 \\
%	\cline{2-5}
	& \small GRADIANT~\cite{boulkenafet2017competition} & \small 2.6$\pm$3.9 & \small 5.0$\pm$5.3 & \small 3.8$\pm$2.4 \\
	\cline{2-5}
	& \small FaceDs~\cite{jourabloo2018face} & \small 4.0$\pm$1.8 & \small 3.8$\pm$1.2 & \small 3.6$\pm$1.6 \\
	\cline{2-5}
	& \small Auxiliary~\cite{liu2018learning} & \small 2.7$\pm$1.3 & \small 3.1$\pm$1.7 & \small 2.9$\pm$1.5 \\
	\cline{2-5}
	& \small STASN~\cite{yang2019face} & \small 4.7$\pm$3.9 & \small \textbf{0.9$\pm$1.2} & \small 2.8$\pm$1.6 \\
	\cline{2-5}
	& \small \textbf{Ours} & \small \textbf{1.5$\pm$1.4} & \small 1.9$\pm$1.9 & \small \textbf{1.7$\pm$1.6} \\
	
	\hline
	
	\multirow{5}{*}{4}
%	& \small MILHP~\cite{lin2018live} & \small 15.8$\pm$12.8 & \small 8.3$\pm$15.7 & \small 12.0$\pm$6.2 \\
%	\cline{2-5}
	& \small GRADIANT~\cite{boulkenafet2017competition} & \small \textbf{5.0$\pm$4.5} & \small 15.0$\pm$7.1 & \small 10.0$\pm$5.0 \\
	\cline{2-5}
	& \small Auxiliary~\cite{liu2018learning} & \small 9.3$\pm$5.6 & \small 10.4$\pm$6.0 & \small 9.5$\pm$6.0 \\
	\cline{2-5}
	& \small STASN~\cite{yang2019face} & \small 6.7$\pm$10.6 & \small 8.3$\pm$8.4 & \small 7.5$\pm$4.7 \\
	\cline{2-5}
	& \small FaceDs~\cite{jourabloo2018face} & \small 5.1$\pm$6.3 & \small 6.1$\pm$5.1 & \small 5.6$\pm$5.7 \\
	\cline{2-5}
	& \small \textbf{Ours} & \small 5.8$\pm$4.9 & \small \textbf{1.7$\pm$2.6} & \small \textbf{3.7$\pm$2.1} \\
	
	\hline
\end{tabular}
\label{intra_oulu}
\end{table}

\subsection{Intra Testing}
We carry out the intra testing on SiW and OULU-NPU datasets. The SiW dataset collects 4478 HD videos from 165 individuals in total. It contains data in the rich PIE (pose, illumination, and expression) variations. There are four pre-defined protocols on the SiW dataset to evaluate the generalization for face anti-spoofing.  The results on the test set are reported in Table~\ref{intra_siw}. The state-of-the-art methods we compare with include Auxiliary \cite{liu2018learning}, STASN \cite{yang2019face} and Meta-FAS-DR \cite{zhao2019meta}. Our method outperforms the state-of-the-art results on all protocols by a large margin. To be notified, protocol 3 evaluates the model's generalization capacity over unknown spoof types and our method obtains considerable performance gain over the state-of-the-art. The experimental results demonstrate the superior generalization capacity of the proposed method. 

The OULU-NPU dataset is a high quality dataset simulating realistic mobile authentication scenarios. It consists of 4950 videos recorded by different mobile phones with front cameras. OULU-NPU develops four protocols to evaluate the model's generalization. Table~\ref{intra_oulu} compares the performance of our model with state-of-the-art methods on the OULU-NPU dataset. The state-of-the-art methods we compare with include GRADIANT~\cite{boulkenafet2017competition}, MILHP~\cite{lin2018live} Auxiliary~\cite{liu2018learning}, STASN~\cite{yang2019face} and FaceDs~\cite{jourabloo2018face}. The experimental results show that our model has great generalization ability over unseen environmental conditions, attack mediums, and camera sensors.

\begin{table}
\caption{The inter testing results between CASIA-MFSD and Replay-Attack datasets and comparison with the state-of-the-art methods. Results under HTER (\%) metric are reported.}
\centering	
\begin{tabular}{|c|c|c|c|c|}
	\hline
	\multirow{3}{*}{Method} & Train & Test & Train & Test \\
	\cline{2-5}
	& \tabincell{c}{CASIA-\\MFSD} & \tabincell{c}{Replay-\\Attack} & \tabincell{c}{Replay-\\Attack} & \tabincell{c}{CASIA-\\MFSD} \\
	\hline
	\small Spectral Cubes~\cite{pinto2015face} & \multicolumn{2}{|c|}{\small 34.4\%} & \multicolumn{2}{|c|}{\small 50.0\%} \\
	\hline
	\small CNN~\cite{yang2014learn} & \multicolumn{2}{|c|}{\small 48.5\%} & \multicolumn{2}{|c|}{\small 45.5\%} \\
	\hline
	\small LBP~\cite{boulkenafet2015face} & \multicolumn{2}{|c|}{\small 47.0\%} & \multicolumn{2}{|c|}{\small 39.6\%} \\ 
	\hline
	\small Color Texture~\cite{boulkenafet2016face} & \multicolumn{2}{|c|}{\small 30.3\%} & \multicolumn{2}{|c|}{\small 37.7\%} \\
	\hline
	\small STASN~\cite{yang2019face} & \multicolumn{2}{|c|}{\small 31.5\%} & \multicolumn{2}{|c|}{\small 30.9\%} \\
	\hline
	\small FaceDs~\cite{jourabloo2018face} & \multicolumn{2}{|c|}{\small 28.5\%} & \multicolumn{2}{|c|}{\small 41.1\%} \\
	\hline
    \small Auxiliary~\cite{liu2018learning} & \multicolumn{2}{|c|}{\small 27.6\%} & \multicolumn{2}{|c|}{\small 28.4\%} \\
	\hline
	\small \textbf{Ours} & \multicolumn{2}{|c|}{\textbf{\small 27.4\%}} & \multicolumn{2}{|c|}{\textbf{\small 23.7\%}} \\
	\hline
	
\end{tabular}
\label{inter_test}
\end{table}

\subsection{Inter Testing}
To demonstrate the generalization capability of our model, we set up inter testing experiments. Specifically, the model is trained on one dataset and then tested on the other dataset. The cross-dataset evaluation is challenging as the data distribution varies a lot on both the live and spoof samples between different datasets. We perform the inter testing on the CASIA-MFSD and Replay-Attack datasets to evaluate the generalization capability of our model. As shown in Table~\ref{inter_test}, our method achieves the best results under the HTER metric among all the state-of-the-art methods. All the state-of-the-art methods perform the inter test on the CASIA-MFSD and Replay-Attack datasets except that Liu \emph{et~al.} \cite{liu2018learning} additionally performs an inter test from SiW to OULU. According to Table~\ref{inter_test} and Table~\ref{inter_test_auxiliary}, the proposed method achieves better performance. To conclude, our method outperforms the state-of-the-art methods on all the conducted inter tests. During the inter test, we notice a slight  performance drop from the high-resolution dataset (CASIA-MFSD) to the low-resolution dataset (Replay-Attack). As demonstrated in Section~\ref{visualizaions}, with high-resolution images as input, the proposed method takes advantage of the abundant texture information, which may be missing on the low-resolution images. On the contrary, the learned spoof cues from low-resolution images can generalize well to high-resolution ones, which is worthy of future research.

\begin{table}[t]
\caption{The inter testing results from SiW to OULU-NPU dataset and comparison with the state-of-the-art methods. Results under ACER (\%) metric are reported.}
\centering
\renewcommand\arraystretch{0.95} 
 \setlength{\tabcolsep}{0.8mm}
\begin{tabular}{|c|c|c|c|c|}
	\hline
	Protocol & \multicolumn{2}{|p{3cm}|}{\centering{Method}} & \multicolumn{2}{|p{3cm}|}{\centering{ACER (\%)}} \\
	\hline
	\multirow{2}{*}{1} &
	  \multicolumn{2}{|c|}{Auxiliary~\cite{liu2018learning}} & \multicolumn{2}{|c|}{10.0} \\
	\cline{2-5}
	& \multicolumn{2}{|c|}{\textbf{Ours}} & \multicolumn{2}{|c|}{\textbf{6.5}} \\
	\hline
	
	\multirow{2}{*}{2} &
	  \multicolumn{2}{|c|}{Auxiliary~\cite{liu2018learning}} & \multicolumn{2}{|c|}{14.1} \\
	\cline{2-5}
	& \multicolumn{2}{|c|}{\textbf{Ours}} & \multicolumn{2}{|c|}{\textbf{11.4}} \\
	\hline
	
	\multirow{2}{*}{3} &
	    \multicolumn{2}{|c|}{Auxiliary~\cite{liu2018learning}} & \multicolumn{2}{|c|}{13.8$\pm$5.7} \\
	\cline{2-5}
	& \multicolumn{2}{|c|}{\textbf{Ours}} &     \multicolumn{2}{|c|}{\textbf{9.9$\pm$8.2}} \\
	\hline
	
	\multirow{2}{*}{4} &
	\multicolumn{2}{|c|}{Auxiliary~\cite{liu2018learning}} & \multicolumn{2}{|c|}{10.0$\pm$8.8} \\
	\cline{2-5}
	& \multicolumn{2}{|c|}{\textbf{Ours}} & \multicolumn{2}{|c|}{\textbf{6.7$\pm$6.6}} \\
	\hline	

\end{tabular}
\label{inter_test_auxiliary}
\end{table}

\begin{table}
\caption{The influence of input image resolution, regression loss (RL), triplet loss (TL) and the auxiliary classification (AC). Results on protocol 1 of OULU-NPU dataset are reported.}
\centering	
\begin{tabular}{p{27pt}p{29pt}p{37pt}p{22pt}p{23pt}p{23pt}p{33pt}p{33pt}|p{30pt}p{30pt}p{30pt}}
	%\hline
	\toprule[1pt]
	\makecell[c]{\scriptsize{Resized}\\\scriptsize{input}} & \makecell[c]{\scriptsize{Patched}\\\scriptsize{input}} & \makecell[c]{\scriptsize{RL on live}\\\scriptsize{and spoof}} &
	\makecell[c]{\scriptsize{RL on}\\\scriptsize{live}} &
	\makecell[c]{\scriptsize{TL on}\\\scriptsize{E5-D4}} &
	\makecell[c]{\scriptsize{TL on}\\\scriptsize{E5-SC}} &
	\makecell[c]{\scriptsize{AC on}\\\scriptsize{spoof cue}} & \makecell[c]{\scriptsize{AC on}\\\scriptsize{overlayed}} & \makecell[c]{\scriptsize{APCER}\\\scriptsize{(\%)}}  & \makecell[c]{\scriptsize{BPCER}\\\scriptsize{(\%)}} & \makecell[c]{\scriptsize{ACER}\\\scriptsize{(\%)}} \\
	%\hline
	\midrule[1pt]
	\makecell[c]{\checkmark} & & & \makecell[c]{\checkmark} & \makecell[c] \checkmark & &  & \makecell[c] \checkmark & \makecell[c]{2.1} & \makecell[c]{3.3} & \makecell[c]{2.7} \\
	\hline
	& \makecell[c] \checkmark & & \makecell[c]{\checkmark} & & & & \makecell[c] \checkmark & \makecell[c]{1.7} & \makecell[c]{3.3} & \makecell[c]{2.5} \\
	\hline
	& \makecell[c] \checkmark & & \makecell[c]{\checkmark} & & \makecell[c] \checkmark & & \makecell[c] \checkmark & \makecell[c]{\textbf{0.2}} & \makecell[c]{4.2} & \makecell[c]{2.3} \\
	\hline
	& \makecell[c] \checkmark & & \makecell[c]{\checkmark} & \makecell[c] \checkmark & & & & \makecell[c]{2.5} & \makecell[c]{0.8} & \makecell[c]{1.7} \\
	\hline
	& \makecell[c] \checkmark & \makecell[c] \checkmark & & \makecell[c] \checkmark & & & \makecell[c] \checkmark & \makecell[c]{0.8} & \makecell[c]{2.5} & \makecell[c]{1.7} \\
	\hline
	& \makecell[c] \checkmark & & \makecell[c]{\checkmark} & \makecell[c] \checkmark & & \makecell[c] \checkmark & & \makecell[c]{1.7} & \makecell[c]{\textbf{0.0}} & \makecell[c]{0.8} \\
	\hline
	& \makecell[c] \checkmark & & \makecell[c]{\checkmark} & \makecell[c] \checkmark & & & \makecell[c] \checkmark & \makecell[c]{0.8} & \makecell[c]{\textbf{0.0}} & \makecell[c]{\textbf{0.4}} \\
	%\hline
	\bottomrule[1pt]
\end{tabular}
\label{ablations_studies}
\end{table}

\subsection{Ablation Studies}
\label{ablations}
\noindent \textbf{Influence of Input Image Resolution.} In our model, we sample patches from images as the network input during training. These patches keep the input image resolution and retain as much texture information as possible. In Table~\ref{ablations_studies}, we compare resizing the cropped faces to 224$\times$224 with the bilinear interpolation and patching 224$\times$224 the patches from the face images. The experimental result shows that patched input strategy achieves better ACER metric. The result is in line with our expectations as our method excavates spoof cues from a single-frame RGB image without extra supervision such as depth or temporal information, the local image information is critical to the performance. On the high resolution dataset like OULU-NPU, the resize operation will blur the image and bring considerable loss in local details, resulting in a performance drop.

\noindent \textbf{Influence of Implicit Supervisions.} In the proposed method we adopt the triplet loss (TL) and the auxiliary classification (AC) in a joint supervision manner. In Table~\ref{ablations_studies}, we remove the triplet learning and the auxiliary classifier separately and report their performances. It shows that the model is most likely to overfitting if only guided by the binary classification supervision. The joint supervision boosts the performance and results in the minimum ACER. To further validate the influence of the triplet loss on the spoof cue map (SC), we set up a comparison experiment which alternatively imposes the triplet loss on feature maps from E5 to SC. The experimental result shows that imposing triplet loss on feature maps from E5 to D4 achieves better performance.

\noindent \textbf{Influence of Regression Loss.} In order to relieve the spoof diversity's influence on the decision boundary modeling, we set explicit regression loss (RL) only for the live samples while setting no regression loss for the spoof ones. To validate the effectiveness of such design, we set up a comparison experiment which imposes regression loss on both live and spoof samples. To be specific, given an RGB image $I$ as input, the spoof cue map $C$ is a zero map for a live sample while an all-one map for a spoof samples. As shown in Table~\ref{ablations_studies}, the proposed design achieves better performance. From our perspective, the predefined explicit regression loss on the spoof samples may violate the intrinsic spoof diversity and makes the network easy to overfitting.

\begin{figure}
\centering
\includegraphics[width=29pc, height=10pc]{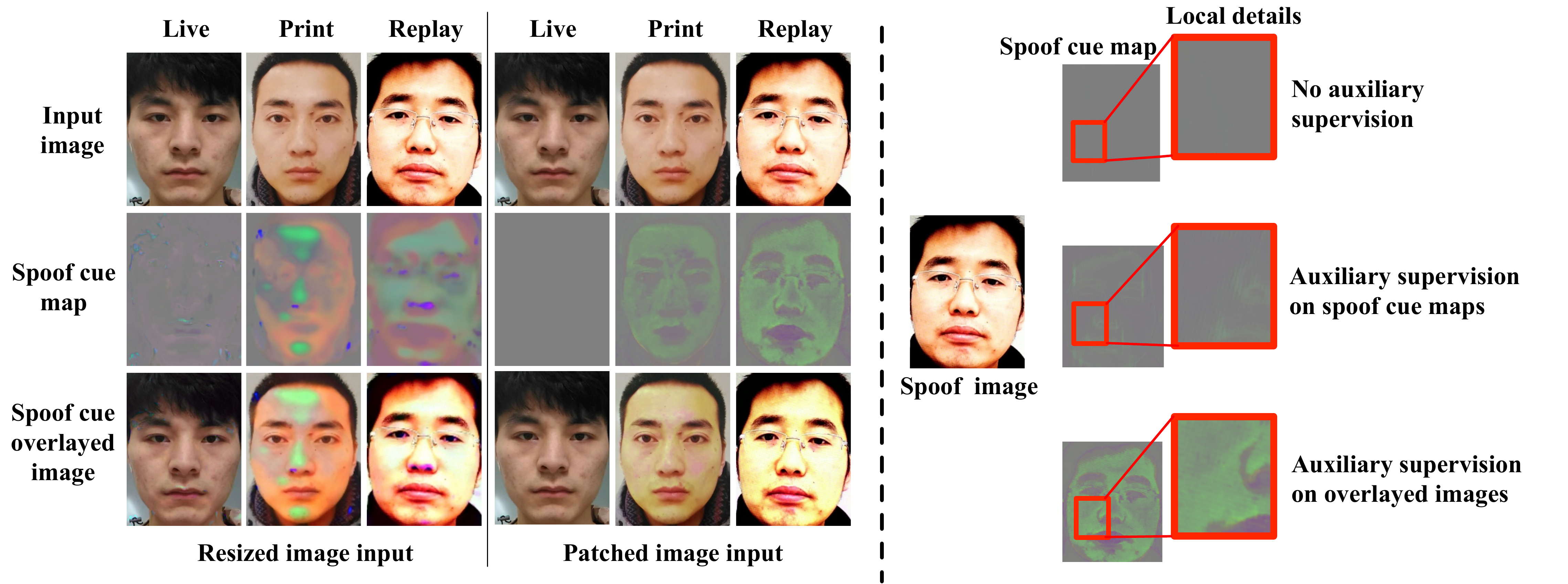}
\caption{The illustration of different types of input and the auxiliary supervision's amplification effect on the spoof cue maps. \textbf{Left}: Results with resized and patched images as input during training. \textbf{Right}: The spoof cue maps and their local details corresponding to different auxiliary supervision strategies.}
\label{aux_on_cue}
\end{figure}

\noindent \textbf{Advantage of Auxiliary Supervision.} In Table~\ref{ablations_studies}, we evaluate the auxiliary classifier and the residual learning approach's influence on the network's performance. The residual learning framework achieves the best ACER performance among all these experiments. For a more intuitive understanding, we present the spoof cue maps from the models in Fig.~\ref{aux_on_cue}. The auxiliary classifier amplifies the spoof cues by enlarging the magnitude of the non-zero values. In our perspective, the correlations between the spoof cue maps and the input images are learned and amplified through the residual learning manner. From the local details column in Fig.~\ref{aux_on_cue}, our method excavates more local spoof details than other auxiliary supervision manners and obtains more discriminative spoof cue representations.

\subsection{Visualizations}
\label{visualizaions}
\noindent \textbf{Visualization of the Spoof Cues.} For a more intuitive understanding of what the spoof cues are, we present the generated spoof cue map in Fig.~\ref{aux_on_cue}. The spoof cues respond to the differences between the live and spoof samples, which could possibly be the color distortion, moire pattern or other spoof patterns. Fig.~\ref{aux_on_cue} shows that the spoof cues exhibit some global pattern when the input image contains a complete face, whereas, for patched input images, the spoof cues contain more local information. In our perspective, the phenomenon is caused by our modeling architecture. Multi-scale skip connections from encoder to decoder and the residual 1$\times$1 Conv in Decoder Residual Block retain image semantics in decoder feature maps well.

\begin{figure}
\centering
\includegraphics[width=15pc, height=11.5pc]{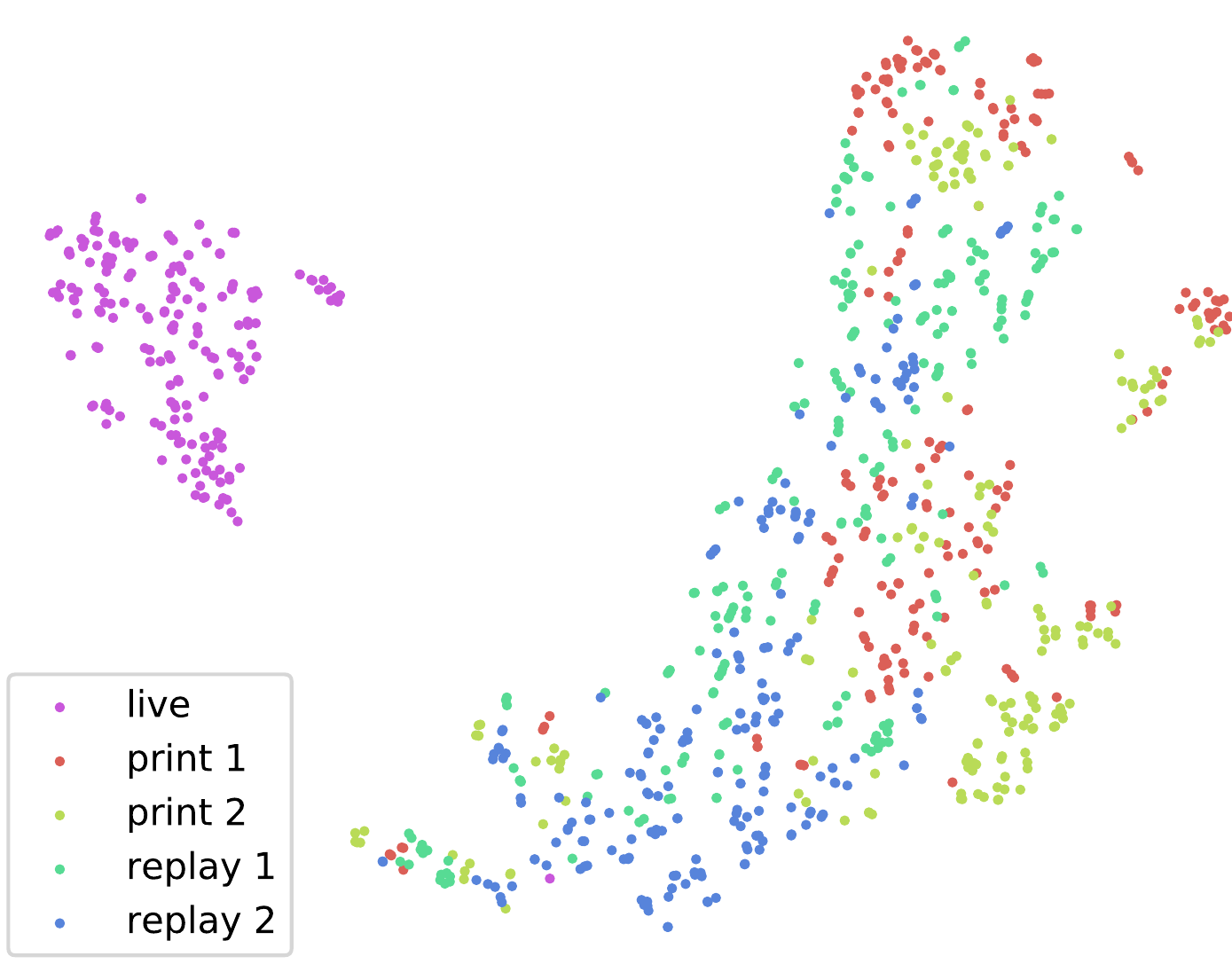}
\caption{The t-SNE representation of feature embeddings of layer D4. The samples for demonstration are chosen from the test set of OULU-NPU on protocol 1.}
\label{tsne}
\end{figure}

\noindent \textbf{Visualization of the Feature Embeddings.} We assume the spoof samples are outliers of the live set and develop the joint supervision method to promote the live-live intra-class compactness and the live-spoof inter-class separability. In Fig.~\ref{tsne}, we present the feature embeddings of layer D4 using t-SNE~\cite{maaten2008visualizing} on the test set of OULU-NPU protocol 1. It shows that all the live samples fall into a cluster while the spoof samples distribute far from this cluster by a considerable margin, which is in line with our desired distribution of the live closed-set and the spoof open-set.

\section{Conclusions}
We reformulate the FAS in an anomaly detection perspective and assume the live samples belong to a closed-set while the spoof samples are outliers from this closed-set belonging to an open-set. We define the spoof cues as the discriminative live-spoof differences and propose a residual learning framework to learn the generalized spoof cues. Our network consists of a spoof cue generator and an auxiliary classifier. In spoof cue generator we impose weak implicit constraint on the spoof samples to guide the proposed network to learn more generalized spoof cues, the auxiliary classifier makes the spoof cues more discriminative and further boosts the spoof cues' generalizability. We conduct extensive experiments on popular datasets and the experimental results show that our method achieves state-of-the-art performance for face anti-spoofing.

\bibliographystyle{splncs04}
\bibliography{egbib}
\end{document}